\documentclass[10pt,twocolumn,letterpaper]{article}

\usepackage{iccv}              

\usepackage{amsmath, amsfonts}
\usepackage{algorithmic}
\usepackage{graphicx}
\usepackage{textcomp}
\usepackage{xcolor}
\usepackage{multirow}
\usepackage{mathtools}
\usepackage{tcolorbox}
\usepackage{enumitem}
\usepackage{makecell}
\usepackage{float}
\usepackage{fancyhdr}
\usepackage{colortbl}
\usepackage{arydshln}
\usepackage{xcolor,soul}
\usepackage{tcolorbox}
\usepackage{wrapfig}
\usepackage{algorithm2e}
\RestyleAlgo{ruled} 
\usepackage{pifont} 
\usepackage{tikz}
\usepackage{booktabs}
\usepackage{listings}

\usepackage[utf8]{inputenc}
\DeclareUnicodeCharacter{2460}{\textcircled{1}}
\DeclareUnicodeCharacter{2461}{\textcircled{2}}
\newcommand{\cmark}{\ding{51}}%
\newcommand{\xmark}{\ding{55}}%
\newcommand\tab[1][1cm]{\hspace*{#1}}
\definecolor{iccvblue}{rgb}{0.21,0.49,0.74}
\usepackage[pagebackref,breaklinks,colorlinks,allcolors=iccvblue]{hyperref}

\def\paperID{9066} 
\def\confName{ICCV}
\def\confYear{2025}

\title{\raisebox{-0.3\height}{\includegraphics[height=10mm]{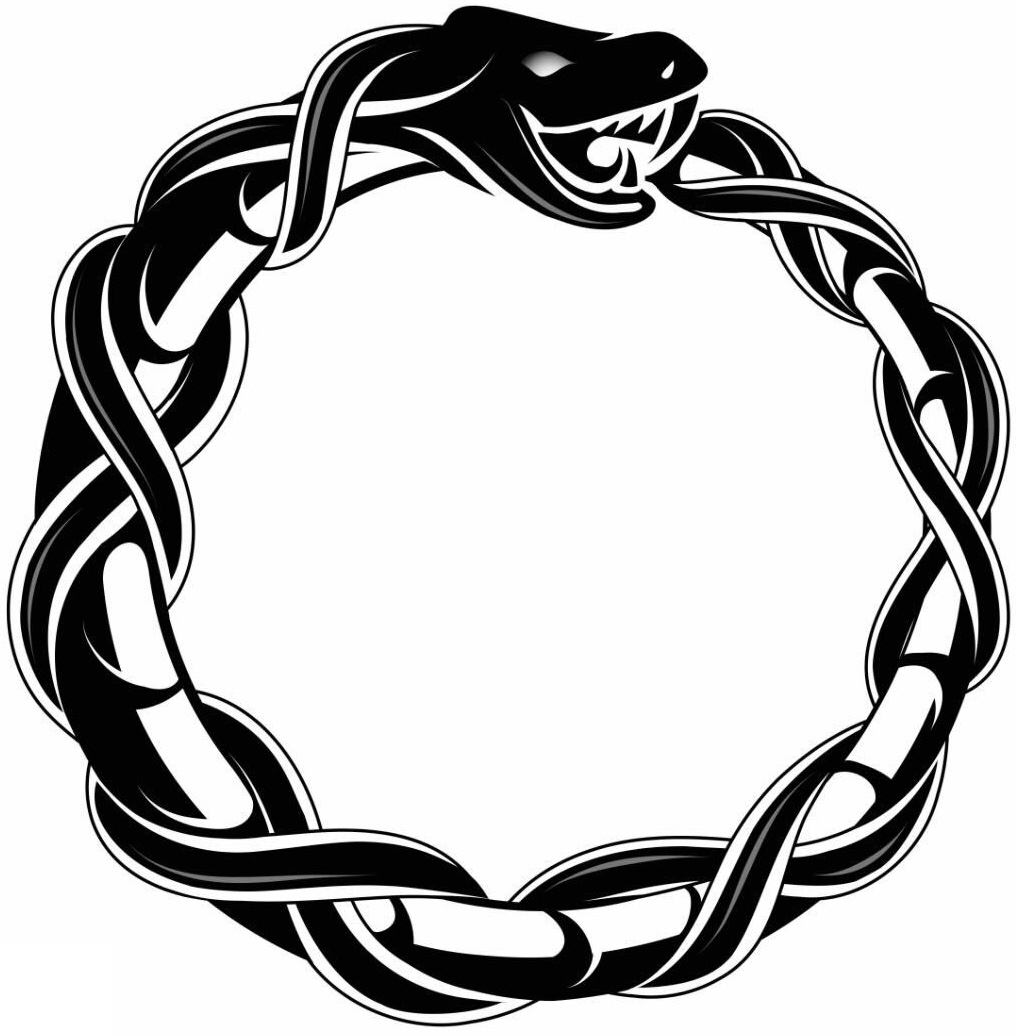}}~OuroMamba: A Data-Free Quantization Framework for Vision Mamba}

\author{Akshat Ramachandran$^{g}{^*}$, Mingyu Lee$^{g}{\thanks{Equal Contribution Authors}}$ , Huan Xu$^{g}$, Souvik Kundu$^{i}$, Tushar Krishna$^{g}$\\
$^{g}$Georgia Institute of Technology, Atlanta, USA  \tab[0.3cm] 
$^{i}$Intel Labs, USA \\
\texttt{\{akshat.r, mlee864, hxu398\}@gatech.edu, tushar@ece.gatech.edu} \\
\texttt{souvikk.kundu@intel.com}
}
\begin{document}

\makeatletter
\let\@oldmaketitle\@maketitle
\renewcommand{\@maketitle}{\@oldmaketitle
    \centering
    \vspace{-7mm}
    \includegraphics[width=0.85\linewidth, keepaspectratio]{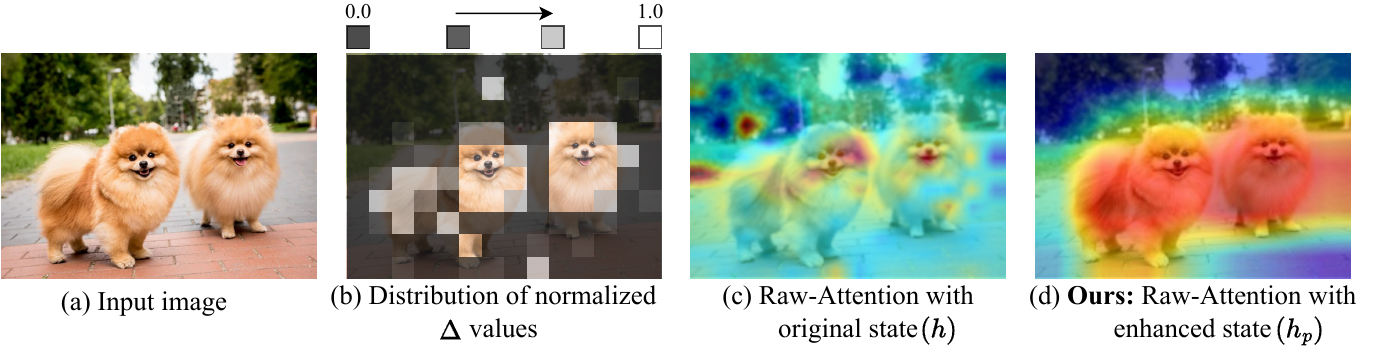}
        \vspace{-4mm}
        \captionof{figure}{Qualitative comparison of VMM implicit attention. (a) Sample input image. (b) Normalized distribution of the input gate values ($\Delta$). Visualization of implicit attention \cite{ali2024hidden} using (c) the original state ($h$) and (d) the \textbf{proposed} patched state ($h_p$), which incorporates spatial dependencies through patched neighborhood interactions.}
        \label{fig:teaser}
        \vspace{1mm}
}%
\makeatother
\maketitle

\begin{abstract}
We present \textbf{OuroMamba}, the first data-free post-training quantization (DFQ) method for vision Mamba-based  models (VMMs). We identify two key challenges in enabling DFQ for VMMs, (1) VMM's recurrent state transitions restricts the capturing of long-range interactions and leads to semantically weak synthetic data, (2) VMM activations exhibit dynamic outlier variations across time-steps, rendering existing static PTQ techniques ineffective. To address these challenges, {OuroMamba} presents a two-stage framework: (1) OuroMamba-\texttt{Gen} to generate semantically rich and meaningful synthetic data. It applies contrastive learning on patch level VMM features generated through neighborhood interactions in the latent state space, (2) OuroMamba-\texttt{Quant} to employ mixed-precision quantization with lightweight dynamic outlier detection. In specific, we present a thresholding based outlier channel selection strategy for activations that is updated every time-step. Extensive experiments across vision and generative tasks show that our \textbf{data-free} OuroMamba surpasses existing data-driven PTQ techniques, achieving state-of-the-art performance across diverse quantization settings. Additionally, we implement efficient GPU kernels to achieve practical latency speedup of up to $\mathbf{2.36}\times$. Code and synthetic dataset are available here: \url{https://github.com/georgia-tech-synergy-lab/ICCV-OuroMamba}.

\end{abstract}
\vspace{-7mm}
\section{Introduction} 
\label{sec:intro}
The sub-quadratic compute alternative of the vision State space models (SSMs) \cite{gu2023mamba}, particularly vision mamba models (VMMs) \cite{zhu2024vision} has made them a promising alternative to ViTs \cite{dosovitskiy2020image}. Like ViTs, larger VMM variants face deployment challenges due to high memory and latency constraints \cite{li2025qmamba}.

Quantization \cite{ramachandran2024clamp, cho2024ptq4vm, yuan2022ptq4vit, ramachandran2025microscopiq} is a popular method to tackle the high memory and latency demand by mapping full-precision (FP) weights and/or activations to lower-bit representations. In particular, post-training quantization (PTQ) \cite{ramachandran2024clamp, kundu2022bmpq, ramachandran2025accelerating} converts a pre-trained FP model to low-precision and requires calibration on a small dataset \cite{ramachandran2024clamp} to recover model performance. However, PTQ's calibration data is typically drawn from the original training dataset, potentially restricting its applicability in situations demanding privacy, security of training set \cite{kundu2021analyzing, zhang2023sal}.

Consequently, data-free quantization (DFQ), a subset of post-training quantization (PTQ) has emerged as a promising alternative. DFQ allows calibration via generating synthetic data from Gaussian noise \cite{ramachandran2024clamp, KimKK25, li2023psaq, li2022patch} mimicking the distribution of the original train-set. 
%
%
In particular, DFQ for ViTs primarily rely on the distinctive response of self-attention to noise and real data to generate synthetic data \cite{li2023psaq, ramachandran2024clamp}. However, despite their significant progress for ViTs, \textit{DFQ techniques for VMM remains largely unexplored}. Interestingly, VMM’s recurrent state transition in the S6 layer \cite{gu2023mamba} lacks an explicit self-attention mechanism \cite{ansari2023gpu, ramachandran2023ntrans}, limiting the usage of any attention based patch similarity improvement. While the S6 layer can be formulated as a variant of linear attention \cite{han2025demystify}, our observations reveal that this implicit attention struggles to distinguish foreground from the background of an image. As observed in \cref{fig:teaser}(c), \cref{fig:gen_motivation}(a), despite VMM's ``theoretical ability" for global interactions \cite{Azizi2025mambaextend}, it fails to capture long-range dependencies in feature representation.           


Notably, VMM’s recurrent state transitions introduce dynamic variations in activation characteristics \cite{li2025qmamba}. These variations occur across time steps and tokens\footnote{We use token and patch interchangeably based on context.}, potentially demanding online adaptive outlier selection. This is in sharp contrast to ViTs, which largely demonstrate a static outlier activation pattern \cite{dong2023packqvit}. A contemporary work, QMamba \cite{li2025qmamba}, rightly identified the dynamic variations in SSM activations. However, it employs statically determined temporal grouped quantization, unable to adapt to the potential changes in outlier locations, yielding large quantization error at low bit precision. Moreover, QMamba lacks a data-free calibration strategy and limited practical speedup, limiting its use in privacy and latency-sensitive scenarios.

\noindent
\textbf{Our Contributions. }To address the above challenges in data generation and quantization stages, we present \textbf{OuroMamba}\footnote{Inspired by \textbf{Ouroboros}—the self-consuming snake— the name OuroMamba reflects its ability to self-generate data for VMM quantization.}, 
a \emph{first-of-its-kind} method to enable DFQ for VMMs. For data generation, we investigate on VMM’s implicit attention struggles to differentiate foreground with background. In specific, we hypothesize it to be dependent on scanning direction of the compressed hidden state ($h(t)$), which limits explicit spatial token interactions. To mitigate this, we propose \textbf{OuroMamba}-\texttt{Gen}, that essentially enhances the implicit attention through patched spatial interactions in the latent state space, resulting in a refined hidden state namely, patched state $h_p(t)$. We then leverage an enhanced implicit attention representation based on $h_p(t)$ (\autoref{fig:teaser}(d)) to generate synthetic data by employing patch-level contrastive learning \cite{ramachandran2024clamp}.


For quantization, we first empirically validate that VMM operation leads to dynamic inter-time-step variations of the outlier channel positions. We then present \textbf{OuroMamba}-\texttt{Quant}, a mixed-precision quantization scheme that performs channel-wise static weight quantization and dynamic activation quantization per-time-step during inference. OuroMamba-\texttt{Quant} reduces the quantization error \cite{lee2025recap} associated to dynamic outliers via three key strategies, (1) a lightweight channel-wise activation outlier detection, (2) mixed-precision quantization, where outlier channels are quantized at higher precision, while remaining inlier channels use lower bit quantization and (3) efficient outlier management using an adaptive outlier list.      

To validate the efficacy of OuroMamba, we conduct extensive experimental evaluations on VMMs \cite{zhu2024vision, liu2025vmamba, huang2024localmamba} and hybrid Transformer-Mamba models \cite{hatamizadeh2024mambavision}  for classification, detection, segmentation, and generative modeling tasks \cite{hu2024zigma}. OuroMamba achieves SoTA performance over existing data-driven VMM PTQ alternatives, with an accuracy improvement of up to 39\%. Additionally, through efficient kernel implementation we demonstrate practical end-to-end latency speedup of up to 2.36$\times$ over FP16 baseline.

\section{Related Works}
\label{sec:related_work}

\noindent
\textbf{Data-Driven PTQ. }Existing PTQ techniques for ViTs \cite{lin2021fq, jiang2025adfq, li2023repq, yang2024dopq, ma2024outlier, yuan2022ptq4vit} address long-tailed distributions and static activation outliers but fail to handle VMMs' dynamically changing outliers \cite{cho2024ptq4vm, li2025qmamba}. While PTQ4VM \cite{cho2024ptq4vm} adapts SmoothQuant \cite{xiao2023smoothquant} to shift activation outliers into weights, it does not quantize SSM activations, limiting its effectiveness. QMamba \cite{li2025qmamba} employs temporal grouped quantization but relies on static groupings, making it unsuitable for ultra-low bit-widths. \emph{OuroMamba is the first to identify dynamic outlier variations in VMMs and introduce a mixed-precision quantization framework that achieves SoTA accuracy at low precision.}

\noindent
\textbf{Date-Free PTQ. }Existing DFQ techniques for ViTs, PSAQ-ViT v1 \cite{li2022patch} and v2 \cite{li2023psaq} optimize noise into synthetic data by maximizing global patch similarity entropy, but overlook spatial sensitivity and semantic inter-patch relationships, leading to simplistic data. CLAMP-ViT \cite{ramachandran2024clamp} addresses this by introducing patch-level contrastive learning \cite{ramachandran2024algorithm} on self-attention, aligning foreground patches while separating background patches. Following data generation, these methods apply uniform symmetric quantization to weights and activations, which is suboptimal for VMM. \emph{DFQ for VMM thus far remains unexplored, and this work makes the first effort in this direction.}

\section{Preliminaries}

\subsection{Mamba Block in VMM}
\label{sec:scan}
\textbf{The S6 Layer. }Each Mamba block in VMM utilizes the selective SSM (S6) mechanism 
\cite{gu2023mamba} that maps an input signal $u(t) \in \mathbb{R}^E$ to an intermediate hidden state $h(t) \in \mathbb{R}^{E \times N}$ through a \textbf{linear recurrent transition}. Here, $E$,$N,t$ correspond to the number of channels, state dimension and time-step, respectively. $h(t)$ then produces the output $o(t) \in \mathbb{R}^E$ via a linear transformation, as shown below.
\vspace{-1mm}
\begin{equation}
    \label{equation:ssm}
    h(t) = \bar{A}(t) \odot h(t-1) + \bar{B}(t) \odot u(t); \hspace{1mm} o(t) = C(t) h(t)
\end{equation}

\noindent
where $\odot$ is element-wise product, \(\bar{A}(t), \bar{B}(t) \in \mathbb{R}^{E \times N}\) and \(C(t) \in \mathbb{R}^{N \times 1}\) are the discrete time-variant system, input, and output matrices, respectively. The S6 layer generates these per-time-step discrete matrices from the input sequence and their corresponding continuous parameter counterparts as follows,
\vspace{-1mm}
\begin{equation}
\label{equation:discretize}
\begin{aligned}
    \bar{A}(t) &= e^{(A \Delta(t))};B(t) = W_B (u(t));
    \bar{B}(t) = B\Delta(t) \\
     \Delta(t) &= \texttt{$S^+$}(u(t)\Delta(t)_{\texttt{proj}});
     C(t) = (W_C(z(t)))^T 
    \end{aligned}
\end{equation}

\noindent
where, $\texttt{$S^+$}$ is softplus, \(\Delta(t)_{\text{proj}}, W_B, \text{ and } W_C\) are linear projection layers. $\Delta(t) \in \mathbb{R}^{E \times 1}$ is the discretization tensor. $\bar{A}, \bar{B}, C, \Delta $ are input-dependent discrete S6 parameters that exhibit dynamic activation variations at each time step. \cref{equation:ssm} deals with timestep-wise inputs. To operate over an input sequence $u$ of batch size $B$, sequence length $M$ with $E$ channels, $u \in \mathbb{R}^{B\times M\times E}$, \cref{equation:discretize}, \cref{equation:ssm} are applied independently to each $u(t) \in \mathbb{R}^{B \times E}$ \cite{gu2023mamba} across $M$ time-steps.

\noindent
\textbf{Mamba Block. }The VMM model is composed by stacking multiple Mamba blocks that maps its input sequence of tokens $X$ to its output sequence $Y$ as follows,
\vspace{-1mm}
\begin{equation}
\begin{aligned}
    G &= \sigma(W_{\textit{gate\_proj}} X); \hspace{0.5mm}
    U = \textit{Conv1D}(W_{\text{in\_proj}} X) \\
    O &= \textit{S6}(U); \hspace{1mm}
    Y = O \odot G
\end{aligned}
\end{equation}

\noindent
here, $G$ is a gating function obtained from a linear transformation of X, followed by a SiLU activation $\sigma$. $\odot$ between $G$ and $O$ enables the model to selectively emphasize or suppress input features. The S6 input, $U$, is a linearly transformed version of $X$, followed by a 1D convolution.

\begin{figure}[t]
  \centering \includegraphics[width = \columnwidth, keepaspectratio]{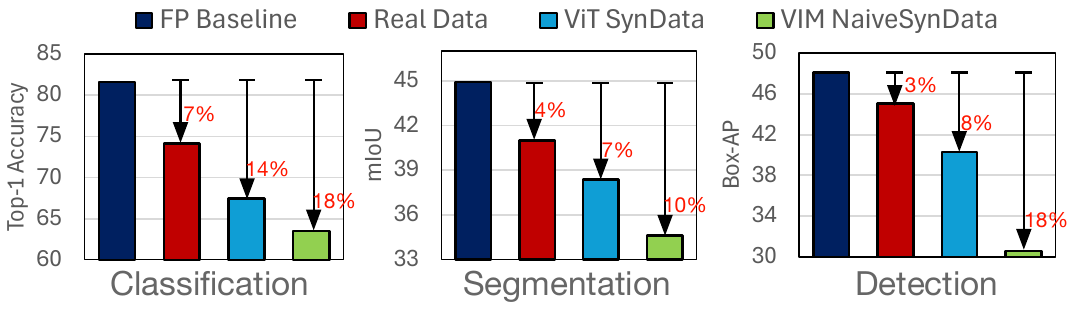}
  \vspace{-8mm}
  \caption{W4A8 quantization performance comparison under the impact of different calibration data sources.}
  \vspace{-6mm}
  \label{fig:gen_motivation2}
\end{figure}

\noindent
\textbf{Selective Scan for Vision. }While selective scan in S6 is well-suited for 1D NLP tasks, vision data is 2D, requiring VMMs to adopt specialized scanning strategies to flatten images into 1D sequences before processing them through Mamba’s selective scan. Examples include 2-way scanning \cite{zhu2024vision} (\cref{fig:gen_motivation2}(a)), 4-way scanning \cite{liu2025vmamba}. However, scan direction choice can introduce directional dependency, impacting the model’s ability to capture spatial relationships \cite{xiao2024grootvl}.

\subsection{Implicit Attention in VMM}
\label{sec:implicit_attention}
Prior works \cite{han2025demystify, ali2024hidden} have shown that S6 layer operation can be reformulated into an implicit self-attention mechanism. By setting $h(0) = 0$ and unrolling \cref{equation:ssm}, the hidden state can be derived as, \( h(t) = \sum_{j=1}^t (\prod_{k=j+1}^{t} \bar{A}(k)) \bar{B}(j) u(j) \) and $o(t)$ can be computed by using this derived value of $h(t)$ in \cref{equation:ssm}. Additionally, in computing $o(t)$, the contribution of the $j^{th}$ token on the $i^{th}$ token is captured as follows,

\vspace{-5mm}
\begin{equation}
\small
\label{equation:attention_score}
   \tilde{\alpha}[i,j] = \sum_{m=1}^N \alpha^m[i,j] = \sum_{m=1}^NC(i)(\prod_{k=j+1}^i\bar{A}(k)_m)\bar{B}(j)[m] 
\end{equation}

\noindent
where, $\bar{A}(k)_m$ is the $m^{\text{th}}$ diagonal element. The output derivation can be simplified as  $o = \tilde{\alpha}u$. Thus, $\tilde{\alpha} \in \mathbb{R}$ is the implicit attention score matrix and $\alpha \in \mathbb{R}^N$ is the $N$-dimensional implicit attention matrix of VMM. 

\subsection{Quantization}
In this paper, we perform symmetric group quantization \cite{ramachandran2024clamp, li2022patch} of both weights and activations for VMMs. The quantization process for an input FP tensor $X$ with a target bit-precision $b$ is given by,
\begin{equation}
\label{equation:quantization}
    Q(X, S, b) ={clip}(\Bigl\lfloor{\frac{X}{S}}\Bigr\rceil, -2^{b-1}+1, 2^{b-1} -1)
\end{equation}

\noindent
where, $S = \texttt{max}(|X|)$ is the scale factor. 

\subsection{Contrastive Objective}
Contrastive learning based on the infoNCE loss \cite{wang2023positive} helps learn an anchor patch from both similar (positive) and dissimilar (negative) patches. Following \cite{ramachandran2024clamp}, the formulation of patch-level contrastive loss ($\mathcal{L}^C_{l,(i,j)}$) we employ for data generation is as follows,
\vspace{-2mm}
\begin{equation}
\scriptsize
\label{equation:contrastive}
\begin{aligned}
    \mathcal{L}^C_{l,(i,j)} = -\log 
    \frac{ \sum_{p+}\exp(\lambda^* \cdot \lambda^{+} / \tau)}
    { \sum_{p+} \exp (\lambda^* \cdot \lambda^{+} / \tau) 
    + \sum_{p-} \exp(\lambda^* \cdot \lambda^{-} / \tau)}
\end{aligned}
\end{equation}
\noindent
where, $*$, $+$, and $-$ correspond to the anchor patch, positive, and negative patches, respectively. $\tau$ controls the concentration level \cite{wang2023positive}. $\lambda_{l,(i,j)}$ (shown without subscript above for brevity) represents the embedding employed to calculate $\mathcal{L}^C_{l,(i,j)}$. $l$ is the layer and $(i,j)$ is the location of anchor patch. The final loss is given as $\mathcal{L}^{C} = \sum_l\sum_i\sum_j\mathcal{L}^{C}_{l,(i,j)}$.
\section{Motivational Analysis}
\label{sec:motivation}

\begin{figure}[t]
  \centering \includegraphics[width = 0.9\columnwidth, keepaspectratio]{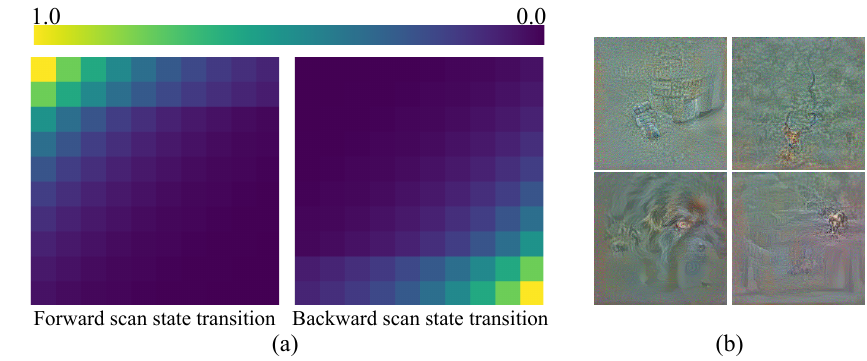}
  \vspace{-4mm}
  \caption{(a) Forward and backward SSM state transition of first layer of Vim-S \cite{zhu2024vision}, (b) Naive synthetic data samples generated by applying \cite{ramachandran2024clamp} on VMM implicit attention.}
  \vspace{-5mm}
  \label{fig:gen_motivation}
\end{figure}

\subsection{Synthetic Data Generation}
\label{sec:motivation1}
\begin{tcolorbox}[colback=gray!10, colframe=black, boxrule=0.5pt, arc=2pt, left=0pt, right=0pt, top=0pt, bottom=0pt]
{\textbf{Observation 1:} \textit{Synthetic data generated for ViT fails to transfer effectively to VMM.}}
\end{tcolorbox}
\noindent
We investigate the W4A8 quantization performance of Vim-S \cite{zhu2024vision} on an existing VMM PTQ framework, QMamba \cite{li2025qmamba}, using different calibration data sources. In specific, we perform calibration with both real data and ViT generated synthetic data of same sample size of 1024 images. We use synthetic calibration data produced by ViT DFQ technique, namely, CLAMP-ViT \cite{ramachandran2024clamp}. \cref{fig:gen_motivation2} reveals that across tasks, ViT synthetic data results in significant accuracy degradation (7-14\%) as compared to using real calibration data.

\begin{figure}[t]
  \centering \includegraphics[width = 0.85\columnwidth, keepaspectratio]{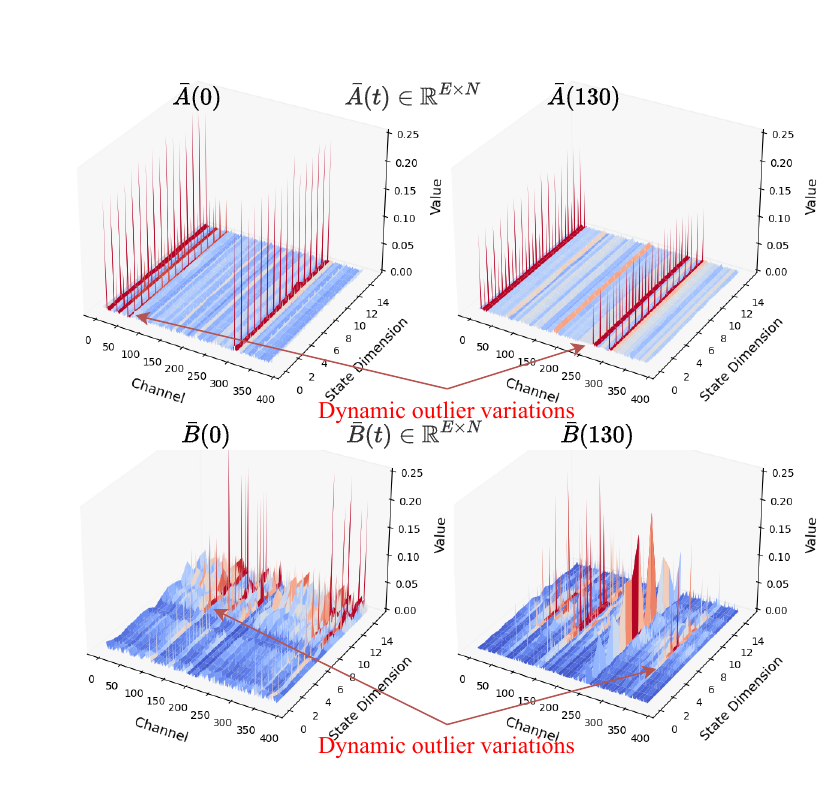}
  \vspace{-2mm}
  \caption{Dynamic inter-time-step outlier channel variations for two representative S6 layer activations: $\bar{A}, \bar{B}$ in layer 3 of Vim-T.}
  \vspace{-5mm}
  \label{fig:outlier_motivation}
\end{figure}

\begin{tcolorbox}[colback=gray!10, colframe=black, boxrule=0.5pt, arc=2pt, left=0pt, right=0pt, top=0pt, bottom=0pt]
{\textbf{Observation 2: }\textit{VMM’s naive implicit attention reformulation is suboptimal for generating synthetic data.}}
\end{tcolorbox}
\noindent
As shown in \cref{fig:teaser}(c), VMM's implicit attention struggles to distinguish foreground from background. This is potentially due to the forgetfulness of the states \cite{Azizi2025mambaextend} as we traverse along the scanning direction (\cref{fig:gen_motivation}(a)) \cite{han2025demystify}. This, unlike self-attention, limits long-range global interactions. Additionally, VMMs \cite{zhu2024vision, liu2025vmamba} employ multiple scanning directions (\cref{sec:scan}) to improve global interactions. However, they process each direction independently with separate S6 blocks. Consequently, each scanning direction’s implicit attention exhibits directional bias and are incoherent \cite{han2025demystify}.

Following \cite{ramachandran2024clamp}, that applies patch-level contrastive loss to ViT self-attention to generate synthetic data, we apply the same method to VMM’s forward scan S6 block and present the generated data in \cref{fig:gen_motivation}(b). The ``naive synthetic data" appears noisy, lacks semantic structure, and performs worse than ViT-generated synthetic data for calibration, \cref{fig:gen_motivation2}). Extending this to other scanning directions yields similar results (see \cref{sec:results}).  
\emph{These observations motivate an optimal VMM data generation technique, capturing spatial dependencies without scanning direction constraints.}

\subsection{Quantization}
\label{sec:motivation2}
\begin{tcolorbox}[colback=gray!10, colframe=black, boxrule=0.5pt, arc=2pt, left=0pt, right=0pt, top=0pt, bottom=0pt]
{\textbf{Observation 3: }\textit{VMM activations exhibit dynamic inter-time-step channel variations.}}
\end{tcolorbox}
\noindent
It has been well established by prior work \cite{dong2023packqvit, jiang2025adfq, yang2024dopq} that {outlier channels in ViT activations follow a fixed pattern}, allowing anticipation in advance during calibration. However, unlike ViTs, \textbf{VMMs exhibit dynamic variations in activation outlier channels} across time-steps. As shown in \cref{fig:outlier_motivation}, a comparison of activations \( \bar{A}, \bar{B} \) at {time-steps 0 and 130} reveals that {outlier channels (in red) differ significantly between time-steps}. This {temporal variance} in outlier positions suggests that a {static, pre-determined calibration approach, as used in ViTs, is insufficient for VMM quantization}.
\emph{Therefore, VMM necessitates dynamic identification of outlier channels at inference to reduce quantization error.}





\section{OuroMamba DFQ Framework}
This section presents OuroMamba, a unified DFQ framework with OuroMamba-\texttt{Gen} for data generation and OuroMamba-\texttt{Quant} for quantization.  

\subsection{Stage 1: OuroMamba-\texttt{Gen}}
\label{sec:gen}
Following the typical DFQ setup \cite{ramachandran2024clamp, li2023psaq}, OuroMamba-\texttt{Gen} requires an input batch of $\mathcal{B}$ random Gaussian noise images $X_{\mathcal{B}}$, and corresponding randomly generated task-specific targets $T_{G_{\mathcal{B}}}$ (detailed in \cref{sec:results}). $X_{\mathcal{B}}$ is fed to the VMM to minimize the generation loss $L^{gen}$ (described below), updating $X_{\mathcal{B}}$ via backpropagation for $G$ iterations. Upon convergence, the updated $X^*_{\mathcal{B}}$ is the generated synthetic data employed for calibration. The algorithm of OuroMamba-\texttt{Gen} pipeline is given in supplementary.  

Building on the observations in \cref{sec:motivation1}, OuroMamba-\texttt{Gen} first constructs an implicit attention representation that possesses enhanced global, spatial interactions for improved synthetic data generation.

\noindent
\textbf{Patched hidden state ($h_p(t)$). }Once the original latent space for a layer is determined, for each time-step of the hidden state $h(t)$ in the original latent space, we define a neighborhood patch $\mathcal{N}(t)$ of size $p \times p$. $\mathcal{N}(t)$ covers all the spatially adjacent tokens $h(k), k \in \mathcal{N}(t)$ in the corresponding 2D representation of the hidden states, where the cardinality of $\mathcal{N}(t)$ denotes the number of neighboring states. For instance, for the $\tau^{th}$ state with $p = 3$, yields total $|\mathcal{N}(\tau)| = 9$ states, with $h(\tau)$ at the center of the patch. For $\tau^{th}$ state, we compute its corresponding \textit{patched state} ($h_p(\tau)$) via a weighted sum of the states in $\mathcal{N}(\tau)$ as, 
\vspace{-1mm}
\begin{equation}
\label{equation:enhanced}
    h_p(\tau) = \sum_{k \in \mathcal{N}(\tau)} w_k h(k)
\end{equation}

\noindent
where, $w_k$ is a weighting factor that modulates the contribution of each neighboring state. Through our experiments and insights from prior work \cite{han2025demystify}, we observe that \( \Delta(t) \) exhibits higher magnitude responses in {informative regions}, such as {foreground}, while suppressing less relevant regions (\cref{fig:teaser}(b)). This property makes \( \Delta(t) \) a {natural choice} for weighting the linear aggregation in \cref{equation:enhanced}, enabling {adaptive feature aggregation}. To leverage this, we determine the weighting factor for all states in $\mathcal{N}(\tau)$ by performing a mean reduction along the channel dimension ($E$) of each \( \Delta(\tau) \),  as given by, $w_k = \textit{mean}_E(\Delta(k))$ for $k \in \mathcal{N}(\tau)$.

\noindent
\textbf{Enhanced Implicit Attention. }By substituting $h(t)$ with $h_p(t)$ in \cref{sec:implicit_attention}, we obtain an enhanced representation of the N-dimensional implicit attention matrix ($\alpha$), denoted as $\alpha_p$. Interestingly, empirical validations show its ability to effectively separate the foreground from background (refer to \cref{fig:teaser}(d)), making it a potentially more informative representation compared to $\alpha$. 

\noindent
\textbf{Generation Loss ($\mathcal{L}^{gen}$). }\cite{ramachandran2024clamp} applied patch-level contrastive loss to self-attention in generating synthetic data for ViT. Inspired by this, we apply the contrastive loss (\cref{equation:contrastive}) on $\alpha_p$. We specifically choose the patch-level contrastive loss as it has been proven by prior work to generate high quality synthetic data through capturing semantic relations. For a VMM layer $l$, each $\alpha^l_p[i,j]$ serves as the anchor patch i.e, $\lambda^*_{l,(i,j)} = \alpha^l_p[i,j]$ with positive and negative patches selected from spatially adjacent $\alpha^l_p[x,y], (x,y) \in  \mathcal{N}(i,j)$. Similar to \cite{ramachandran2024clamp}, we leverage the cosine similarity metric to identify positive and negative patches in $\mathcal{N}(i,j)$, where we select the top-$n$ patches in $\mathcal{N}(i,j)$ with highest similarity as positive and rest as negative, where $n = \lfloor |\mathcal{N}(i,j)|/2\rfloor$. We then compute $\mathcal{L}^C_{l,(i,j)}$ for all anchor patches in $\alpha^l_p$ across all layers to get the final contrastive learning objective $\mathcal{L}^C$.

Furthermore, to direct the synthetic data generation process towards task-specific goals \cite{li2023psaq, ramachandran2024clamp}, we employ an additional output loss $\mathcal{L}^O$, which is the mean absolute error (MAE) between the predicted output with the task specific targets. The final generation loss is thus $L^{gen}=\mathcal{L}^C + \mathcal{L}^O$.

\subsection{Stage 2: OuroMamba-\texttt{Quant}}
\label{sec:quant}
Building on the insights from \cref{sec:motivation2}, OuroMamba-\texttt{Quant} is designed to accommodate the dynamic nature of VMMs. OuroMamba-\texttt{Quant}'s algorithm for activation quantization per-time-step is given in supplementary.

\noindent
\textbf{Offline Calibration. }We use synthetic data generated by OuroMamba-\texttt{Gen} for calibration. \textbf{During calibration}, for an input activation  tensor $X(t)$, we identify the per-time-step inlier scale factor \( S^I(t) \) shared over each tensor of size $N \times E$. We then compute the magnitude-based threshold $\theta$ that distinguishes outliers by identifying values exceeding this threshold. While outliers exhibit inter-channel variations across time steps, their magnitude and dynamic range remain consistent, allowing a static determination of $\theta$.   

\noindent
\textbf{Dynamic Outlier Detection. }To dynamically detect outlier channels \textbf{during inference} at each time step, we employ an adaptive selection mechanism alongside an outlier list ($O_{\texttt{list}}$) initialized to \texttt{NULL} that tracks outlier channels. At each time-step $t$, we analyze the activation tensor \textit{online} and compute a dynamic scale factor $S^D(t)$ of the whole tensor not including channels in $O_{\texttt{list}}$. For $S^D(t) \leq S^I(t)$, it indicates that no new outlier channels are present. However, if $S^D(t) > S^I(t)$, the increase in scale factor suggests the presence of high magnitude outliers. In this case, we iterate through all activation channels, and compute the per-channel scale factor over each channel of size $N\times1$. We then compare the per-channel scale factor with the threshold $\theta$ to identify outliers channels. The identified outlier channels are added to $O_{\texttt{list}}$. $O_{\texttt{list}}$ is propagated across time-steps accumulating new outlier channels at every time-step while also retaining outlier channels of previous time-steps. However, such an approach may result in outdated or transient outlier channels being retained. To prevent this, we introduce a periodic refresh mechanism that updates the $O_{\texttt{list}}$ every $n_\texttt{refresh}$ time-steps to \texttt{NULL}. 

\noindent
\textbf{Mixed Precision Quantization. } To prevent significant accuracy degradation, we keep dynamically identified outliers at higher precision as compared to the inliers. Therefore, at every time-step each outlier channel having dedicated scale factor, are symmetric quantized to $b_a^O$-bits via per-channel quantization. On the other hand, inliers having a shared scale factor $S^I(t)$ over the tensor, are symmetric group quantized to $b_a^I$-bits ($b_a^I < b_a^O$) following \cref{equation:quantization}.

\noindent
\textbf{Weight Quantization. } We apply per-channel symmetric group quantization for the weights. In specific, we follow \cref{equation:quantization} to apply $b_w$-bit  per-channel static quantization, where each channel $c$ has a shared scale factor $S^W_c$.

\section{Experimental Results}
\label{sec:results}

\begin{figure}[t]
  \centering \includegraphics[width = \columnwidth, keepaspectratio]{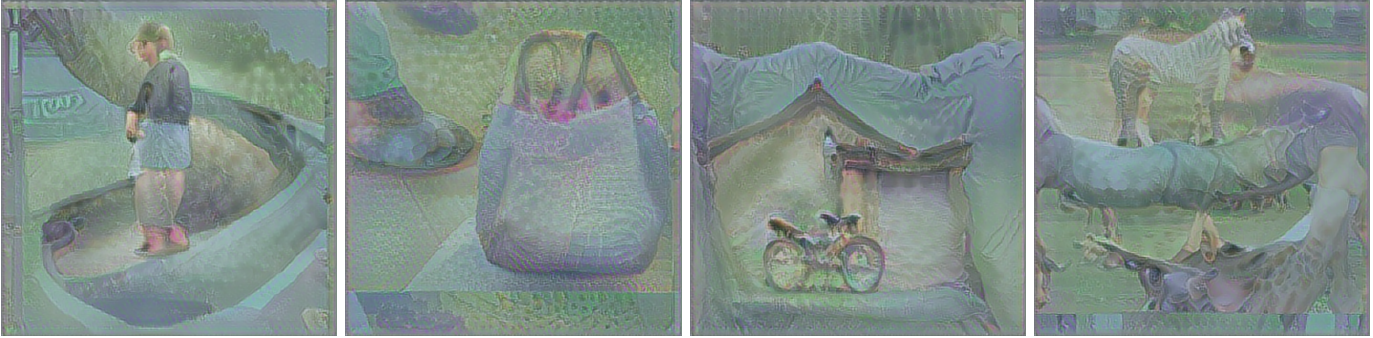}
  \vspace{-5mm}
  \caption{Synthetic data samples generated by OuroMamba.}
  \vspace{-5mm}
  \label{fig:synthetic_samples}
\end{figure}

\subsection{Experimental Setup} 
\textbf{Models and Datasets. }We evaluate OuroMamba across a range of VMM and hybrid Transformer-Mamba models, covering fundamental CV tasks including classification, detection, segmentation, and image generation tasks, as detailed below. Please see supplementary on details of extension of OuroMamba-Quant to transformer layers.

\noindent
\textit{Image Classification. }We employ ImageNet-1K \cite{deng2009imagenet} with 50K tests for ViM-T/S/B \cite{zhu2024vision}, VMamba-T/B \cite{liu2025vmamba}, LVMamba-S \cite{huang2024localmamba} and hybrid MambaVision-T/B \cite{hatamizadeh2024mambavision}.

\noindent
\textit{Object Detection. }We use the COCO 2017 dataset \cite{lin2015coco} having 20K test data. Following \cite{ramachandran2024clamp, li2023psaq}, we use the Mask R-CNN \cite{he2017mask} framework with VMamba-S \cite{liu2025vmamba} and MambaVision-T \cite{hatamizadeh2024mambavision} as the backbones.

\noindent
\textit{Semantic Segmentation. }We avail the ADE20K dataset \cite{zhou2019semantic} with 3K test data encompassing 150 categories with VMamba-S \cite{liu2025vmamba} and MambaVision-T \cite{hatamizadeh2024mambavision} as the backbones. We employ the UperNet framework \cite{xiao2018unified}.

\noindent
\textit{Image Generation. }We evaluate performance using two datasets FacesHQ \cite{FFHQ_NVIDIA} and LandscapesHQ \cite{2021landscape} on Zigma, a zigzag Mamba backbone based diffusion model \cite{hu2024zigma}.


\noindent
\textbf{Baselines for Comparison. }We compare OuroMamba with state-of-the-art (SoTA) VMM PTQ techniques, PTQ4VM \cite{cho2024ptq4vm} and QMamba \cite{li2025qmamba}. Additionally, we compare with ViT PTQ methods, BREC-Q \cite{li2021brecq} and DopQ-ViT \cite{yang2024dopq}, applying them to VMMs for a comprehensive evaluation.

\noindent
\textbf{Implementation Details. }OuroMamba is implemented in PyTorch. All experiments are conducted on a single NVIDIA A100 GPU. Please see \autoref{tab:hyperparameters} for different hyperparameter values used in the evaluations.

\noindent
\textbf{Task-Specific Synthetic Data Generation. }For calibration, we use $\mathcal{B} = 128$ synthetic samples. Following \cite{ramachandran2024clamp}, for image classification on the ImageNet-1K, we create $T_{G_{\mathcal{B}}} \in \mathcal{R}^{\mathcal{B} \times 1000}$, where the class-wise probabilities are randomly determined and assigned. Similarly, the target for object detection is $T_{G_{\mathcal{B}}} \in \mathcal{R}^{\mathcal{B} \times bb \times 5}$ where $bb$ is the number of bounding boxes in the image that is randomly selected from the integer set $[1,3]$. $T_{G_{\mathcal{B}}}[\mathcal{B}, :, 0]$ corresponds to the bounding box category and $T_{G_{\mathcal{B}}}[\mathcal{B}, :, 1:4]$ is the bounding box coordinates $x,y,w,h$ \cite{huang2018introduction}. For segmentation, the target is a pixel-wise classification map of the same size as $X_{\mathcal{B}}$.      

\noindent
\textbf{Quantization Setting Notation. }Following prior work \cite{lin2024qserve, 2024mixq, ramachandran2024algorithm}, the notation for a mixed-precision scheme such as OuroMamba is, W($b_w$)A($b_a^I$)O($b_a^O$). Since, in our evaluations we fix $b_a^O = 8$, we omit $O8$ from the results for brevity.  
\begingroup  
\begin{table}[t]\centering
 \caption{Quantization accuracy comparison for ImageNet classification. `R', `S' signifies real and synthetic calibration data.}
 \vspace{-10pt}
  \renewcommand*{\arraystretch}{1.0}
  \setlength\tabcolsep{1.9pt}
\resizebox{\linewidth}{!}
{%
\begin{tabular}{clcc|cc|cc}
\Xhline{2\arrayrulewidth}
Model & Method & Data & \#Images &  W/A & Top-1 & W/A & Top-1 \\
\Xhline{2\arrayrulewidth}
\multirow{6}{*}{Vim-S \cite{zhu2024vision}} 
& \cellcolor[HTML]{D3D3D3}FP Baseline 
& \cellcolor[HTML]{D3D3D3}- 
& \cellcolor[HTML]{D3D3D3}- 
& \cellcolor[HTML]{D3D3D3}32/32 
& \cellcolor[HTML]{D3D3D3}81.60 
& \cellcolor[HTML]{D3D3D3}32/32 
& \cellcolor[HTML]{D3D3D3}81.60 \\ \cline{2-8}

& BRECQ \cite{li2021brecq} & R & 1024 & 4/8 & 57.32 & 4/4 & 22.46 \\
& DopQ-ViT \cite{yang2024dopq} & R & 1024 & 4/8 & 68.96 & 4/4 & 61.58 \\
& PTQ4VM \cite{cho2024ptq4vm} & R & 256 & 4/8 & 74.37 & 4/4  & 69.60 \\
& QMamba \cite{li2025qmamba} & R & 1024 &  4/8 & 74.12 & 4/4  & 33.64 \\

& \cellcolor[HTML]{D5E8D4}\textbf{OuroMamba (Ours)} 
& \cellcolor[HTML]{D5E8D4}S 
& \cellcolor[HTML]{D5E8D4}128 
& \cellcolor[HTML]{D5E8D4}4/8 
& \cellcolor[HTML]{D5E8D4}\textbf{\underline{79.81}} 
& \cellcolor[HTML]{D5E8D4}4/4 
& \cellcolor[HTML]{D5E8D4}\textbf{\underline{75.93}} \\
 \Xhline{1\arrayrulewidth}

\multirow{6}{*}{Vim-B \cite{zhu2024vision}} 
& \cellcolor[HTML]{D3D3D3}Baseline 
& \cellcolor[HTML]{D3D3D3}- 
& \cellcolor[HTML]{D3D3D3}- 
& \cellcolor[HTML]{D3D3D3}32/32 
& \cellcolor[HTML]{D3D3D3}81.90 
& \cellcolor[HTML]{D3D3D3}32/32 
& \cellcolor[HTML]{D3D3D3}81.90 \\ \cline{2-8}

& BRECQ \cite{li2021brecq} & R & 1024 & 4/8 & 61.52 & 4/4 & 27.34 \\
& DopQ-ViT \cite{yang2024dopq} & R & 1024 & 4/8 & 68.47 & 4/4 & 62.33 \\
& PTQ4VM \cite{cho2024ptq4vm} & R & 256 & 4/8 & 71.02 & 4/4  & 55.60 \\
& QMamba \cite{li2025qmamba} & R & 1024 &  4/8 & 75.46 & 4/4  & 66.73 \\

& \cellcolor[HTML]{D5E8D4}\textbf{OuroMamba (Ours)} 
& \cellcolor[HTML]{D5E8D4}S 
& \cellcolor[HTML]{D5E8D4}128 
& \cellcolor[HTML]{D5E8D4}4/8 
& \cellcolor[HTML]{D5E8D4}\textbf{\underline{80.17}} 
& \cellcolor[HTML]{D5E8D4}4/4 
& \cellcolor[HTML]{D5E8D4}\textbf{\underline{77.34}} \\

\Xhline{1\arrayrulewidth}  
\noalign{\vspace{2pt}}  
\Xhline{1\arrayrulewidth} 

\multirow{6}{*}{VMamba-B \cite{liu2025vmamba}} 
& \cellcolor[HTML]{D3D3D3}FP Baseline 
& \cellcolor[HTML]{D3D3D3}- 
& \cellcolor[HTML]{D3D3D3}- 
& \cellcolor[HTML]{D3D3D3}32/32 
& \cellcolor[HTML]{D3D3D3}83.90 
& \cellcolor[HTML]{D3D3D3}32/32 
& \cellcolor[HTML]{D3D3D3}83.90 \\ \cline{2-8}

& BRECQ \cite{li2021brecq} & R & 1024 & 4/8 & 62.34 & 4/4 & 25.65 \\
& DopQ-ViT \cite{yang2024dopq} & R & 1024 & 4/8 & 68.12 & 4/4 & 61.48 \\
& PTQ4VM \cite{cho2024ptq4vm} & R & 256 & 4/8 & 78.95 & 4/4  & 75.67 \\
& QMamba \cite{li2025qmamba} & R & 1024 &  4/8 & 76.12 & 4/4  & 59.35 \\

& \cellcolor[HTML]{D5E8D4}\textbf{OuroMamba (Ours)} 
& \cellcolor[HTML]{D5E8D4}S 
& \cellcolor[HTML]{D5E8D4}128 
& \cellcolor[HTML]{D5E8D4}4/8 
& \cellcolor[HTML]{D5E8D4}\textbf{\underline{82.03}} 
& \cellcolor[HTML]{D5E8D4}4/4 
& \cellcolor[HTML]{D5E8D4}\textbf{\underline{78.91}} \\

\Xhline{1\arrayrulewidth}  
\noalign{\vspace{2pt}}  
\Xhline{1\arrayrulewidth} 

\multirow{6}{*}{LVMamba-S \cite{huang2024localmamba}} 
& \cellcolor[HTML]{D3D3D3}FP Baseline 
& \cellcolor[HTML]{D3D3D3}- 
& \cellcolor[HTML]{D3D3D3}- 
& \cellcolor[HTML]{D3D3D3}32/32 
& \cellcolor[HTML]{D3D3D3}83.70 
& \cellcolor[HTML]{D3D3D3}32/32 
& \cellcolor[HTML]{D3D3D3}83.70 \\ \cline{2-8}

& BRECQ \cite{li2021brecq} & R & 1024 & 4/8 & 55.34 & 4/4 & 11.62 \\
& DopQ-ViT \cite{yang2024dopq} & R & 1024 & 4/8 & 61.37 & 4/4 & 54.96 \\
& PTQ4VM \cite{cho2024ptq4vm} & R & 256 & 4/8 & 81.46 & 4/4  & 78.21 \\
& QMamba \cite{li2025qmamba} & R & 1024 &  4/8 & 75.19 & 4/4  & 64.24 \\

& \cellcolor[HTML]{D5E8D4}\textbf{OuroMamba (Ours)} 
& \cellcolor[HTML]{D5E8D4}S 
& \cellcolor[HTML]{D5E8D4}128 
& \cellcolor[HTML]{D5E8D4}4/8 
& \cellcolor[HTML]{D5E8D4}\textbf{\underline{82.94}} 
& \cellcolor[HTML]{D5E8D4}4/4 
& \cellcolor[HTML]{D5E8D4}\textbf{\underline{80.11}} \\

\Xhline{1\arrayrulewidth}  
\noalign{\vspace{2pt}}  
\Xhline{1\arrayrulewidth} 

\multirow{6}{*}{\shortstack{\textbf{Hybrid Model} \\ MambaVision-B \cite{hatamizadeh2024mambavision}}} 
& \cellcolor[HTML]{D3D3D3}FP Baseline 
& \cellcolor[HTML]{D3D3D3}- 
& \cellcolor[HTML]{D3D3D3}- 
& \cellcolor[HTML]{D3D3D3}32/32 
& \cellcolor[HTML]{D3D3D3}84.20 
& \cellcolor[HTML]{D3D3D3}32/32 
& \cellcolor[HTML]{D3D3D3}84.20 \\ \cline{2-8}

& BRECQ \cite{li2021brecq} & R & 1024 & 4/8 & 64.59 & 4/4 & 47.38 \\
& DopQ-ViT \cite{yang2024dopq} & R & 1024 & 4/8 & 70.68 & 4/4 & 61.43 \\
& PTQ4VM \cite{cho2024ptq4vm} & R & 256 & 4/8 & 76.51 & 4/4  & 71.27 \\
& QMamba \cite{li2025qmamba} & R & 1024 &  4/8 & 76.08 & 4/4  & 68.59 \\

& \cellcolor[HTML]{D5E8D4}\textbf{OuroMamba (Ours)} 
& \cellcolor[HTML]{D5E8D4}S 
& \cellcolor[HTML]{D5E8D4}128 
& \cellcolor[HTML]{D5E8D4}4/8 
& \cellcolor[HTML]{D5E8D4}\textbf{\underline{82.97}} 
& \cellcolor[HTML]{D5E8D4}4/4 
& \cellcolor[HTML]{D5E8D4}\textbf{\underline{79.24}} \\
 
 \Xhline{2\arrayrulewidth}
\end{tabular}
}
\vspace{-3mm}
\label{tab:classification}
\end{table}
\endgroup
\vspace{-2mm}
\subsection{Analysis of Generated Synthetic Data}
In \cref{fig:synthetic_samples}, we visualize the synthetic samples generated by OuroMamba-\texttt{Gen} for Vim-B \cite{zhu2024vision} for image classification (refer supplementary for additional visualizations). Evidently, OuroMamba-\texttt{Gen} generates, realistic and clear class-specific foreground objects in contextually suitable background, showcasing a sophisticated understanding of semantic relationships between patches. This can be attributed to proposed enhanced impliict attention representation that is able to capture complex spatial relationships. This is in stark contrast to the naive synthetic samples in \cref{fig:gen_motivation}(b), which are noisy and lack semantic structure.

\begingroup	
\begin{table}[t]\centering
 \caption{Quantization performance comparison for detection on COCO 2017. `R', `S' signifies real and synthetic calibration data.}
 \vspace{-10pt}
  \renewcommand*{\arraystretch}{1.0}
  \setlength\tabcolsep{1.9pt}
\resizebox{\linewidth}{!}
{%
\begin{tabular}{lc|ccc|ccc}
\Xhline{2\arrayrulewidth}
& & \multicolumn{3}{c|}{VMamba-S \cite{liu2025vmamba}} & \multicolumn{3}{c}{MambaVision-T \cite{hatamizadeh2024mambavision}}  \\
\Xhline{1\arrayrulewidth}
 Method & Data & W/A & AP$^{box}$ & AP$^{mask}$ & W/A & AP$^{box}$ & AP$^{mask}$  \\
\Xhline{2\arrayrulewidth}
\rowcolor[HTML]{D3D3D3}
Baseline & - & 32/32 & 48.7 & 43.7 & 32/32 & 46.4 & 41.8 \\
  \Xhline{1\arrayrulewidth}

 BREC-Q \cite{li2021brecq} & R & 4/8 & 35.6 & 33.7 & 4/8 & 39.9 & 40.1 \\
 DopQ-ViT \cite{yang2024dopq} & R & 4/8 & 39.2 & 39.1 & 4/8 & 40.8 & 40.5 \\
 PTQ4VM \cite{cho2024ptq4vm} & R & 4/8 & \underline{\textbf{48.4}} & 42.0 & 4/8 & \underline{\textbf{46.2}} & 40.9 \\
 QMamba \cite{li2025qmamba} & R & 4/8 & 47.2 & 41.3 & 4/8 & 45.3 & 39.9 \\
  \rowcolor[HTML]{D5E8D4}
  \textbf{OuroMamba (Ours)} & S & 4/8 & 48.3  & \underline{\textbf{42.9}} & 4/8 & 46.1  & \underline{\textbf{41.4}} \\
 \Xhline{1\arrayrulewidth}
 BREC-Q \cite{li2021brecq} & R & 4/4 & 27.3 & 25.4 & 4/4 & 29.6 & 25.1 \\
 DopQ-ViT \cite{yang2024dopq} & R & 4/4 & 31.5 & 31.1 & 4/4 & 30.9 & 29.7 \\
 PTQ4VM \cite{cho2024ptq4vm} & R & 4/4 & 45.6 & 41.3 & 4/4 & 43.8 & 39.4 \\
 QMamba \cite{li2025qmamba} & R & 4/4 & 43.7 & 39.1 & 4/4 & 42.1 & 38.6 \\
  \rowcolor[HTML]{D5E8D4}  
  \textbf{OuroMamba (Ours)} & S & 4/4 & \underline{\textbf{47.8}}  & \underline{\textbf{42.5}} & 4/4 & \underline{\textbf{44.9}}  & \underline{\textbf{40.9}} \\
 \Xhline{2\arrayrulewidth}
\end{tabular}
}
\vspace{-4mm}
\label{tab:object}
\end{table}
\endgroup
\begingroup	
\begin{table}[t]\centering
 \caption{Quantization performance comparison for segmentation on ADE20K. `R', `S' signifies real and synthetic calibration data.}
 \vspace{-6pt}
  \renewcommand*{\arraystretch}{1.0}
  \setlength\tabcolsep{1.9pt}
\resizebox{\linewidth}{!}
{%
\begin{tabular}{lc|cc|cc}
\Xhline{2\arrayrulewidth}
& & \multicolumn{2}{c|}{VMamba-S \cite{liu2025vmamba}} & \multicolumn{2}{c}{MambaVision-T \cite{hatamizadeh2024mambavision}}  \\
\Xhline{1\arrayrulewidth}
 Method & Data & W/A & mIoU & W/A & mIoU  \\
\Xhline{2\arrayrulewidth}
\rowcolor[HTML]{D3D3D3}
Baseline & - & 32/32 & 50.6 & 32/32 & 46.6 \\
  \Xhline{1\arrayrulewidth}
  PTQ4VM \cite{cho2024ptq4vm} & R & 4/4 & 42.6 & 4/4 & 40.6 \\
  QMamba \cite{li2025qmamba} & R & 4/4 & 40.7 & 4/4 & 38.5 \\

\rowcolor[HTML]{D5E8D4}  
  \textbf{OuroMamba (Ours)} & S & 4/4 & \underline{\textbf{47.3}} & 4/4 & \underline{\textbf{44.1}} \\
 \Xhline{2\arrayrulewidth}
\end{tabular}
}
\vspace{-4mm}
\label{tab:semantic}
\end{table}
\endgroup
\vspace{-1mm}
\subsection{Quantization Results for Classification}
\label{sec:classification}
As shown in \autoref{tab:classification}, existing ViT PTQ techniques \cite{yang2024dopq, li2021brecq} suffer from significant accuracy degradations--particularly at W4A4--of up to $72\%$ on VMM models and up to $37\%$ on hybrid models. Clearly demonstrating the limitations of ViT PTQ methods in handling the dynamic outlier characteristics of VMMs. On the other hand, baseline VMM PTQ methods PTQ4VM \cite{cho2024ptq4vm}, QMamba \cite{li2025qmamba} achieve acceptable performance at W4A8. However, at W4A4, they face large accuracy drops of up to $47\%$, due to static temporal grouping, per-token static scale factors. Furthermore, PTQ4VM's migration of outliers from activations to weights, complicates weight quantization, contributing to its accuracy degradation. In contrast, OuroMamba consistently outperforms baselines across different models and W/A settings, achieving near-lossless performance, with an accuracy drop of $\leq 1.8\%$ at W4A8. Similarly, at W4A4 OuroMamba yields an \textbf{accuracy improvement of} $\mathbf{7.84}\%$ \textbf{over PTQ4VM and $\mathbf{19.40}\%$ over QMamba on average, at W4A4}. It is important to note that, OuroMamba's superior performance is achieved in a data-free scenario requiring only 128 synthetic data samples, compared to PTQ4VM and QMamba requiring 256 and 1024 real calibration data.


\subsection{Quantization Results for Object Detection}
\autoref{tab:object} shows the quantization performance of OuroMamba and the baselines for object detection. Across different quantization settings, OuroMamba consistently yield superior performance yielding up to $\mathbf{21.1}$ and $\mathbf{18.1}$ \textbf{higher box AP and mask AP}, respectively.

\subsection{Quantization Results for Segmentation}
We present our evaluation for segmentation in \autoref{tab:semantic}. OuroMamba at W4A4 achieves upto $\mathbf{6.6}$ \textbf{higher mIoU as compared to PTQ4VM and QMamba}. 

\begingroup	
\begin{table}[t]\centering
 \caption{Zigma image generation fidelity comparison under different quantization techniques.}
 \vspace{-6pt}
  \renewcommand*{\arraystretch}{1.0}
  \setlength\tabcolsep{1.9pt}
\resizebox{0.8\linewidth}{!}
{%
\begin{tabular}{l|cc|cc}
\Xhline{2\arrayrulewidth}
& \multicolumn{2}{c|}{Faces HQ \cite{FFHQ_NVIDIA}} & \multicolumn{2}{c}{Landscape HQ \cite{skorokhodov2021aligning}}  \\
\Xhline{1\arrayrulewidth}
Method & W/A & FID ($\downarrow$) & W/A & FID ($\downarrow$) \\ \Xhline{2\arrayrulewidth}
\rowcolor[HTML]{D3D3D3}
Baseline & 16/16 & 37.8 & 16/16 & 10.7 \\
  \Xhline{1\arrayrulewidth}
  PTQ4VM \cite{cho2024ptq4vm} & 4/4 & 89.6 & 4/4 & 46.2 \\
  QMamba \cite{li2025qmamba} & 4/4 & 90.3 & 4/4 & 41.3 \\

\rowcolor[HTML]{D5E8D4}  
  \textbf{OuroMamba (Ours)} & 4/4 & \underline{\textbf{39.2}} & 4/4 & \underline{\textbf{11.1}} \\
 \Xhline{2\arrayrulewidth}
\end{tabular}
}
\vspace{-10pt}
\label{tab:diffusion}
\end{table}
\endgroup

\subsection{Quantization Results for Diffusion Models}
\noindent
\textbf{Quantitative Analysis. }Following the experimental setup in \cite{hu2024zigma}, we assess the image generation fidelity of Zigma \cite{hu2024zigma}, a VMM-based diffusion model, under W4A4 quantization. Fidelity is measured using the Fréchet Inception Distance (FID) score \cite{heusel2017gans} across 5k samples, with results summarized in \autoref{tab:diffusion}. Notably, across both evaluated datasets, the \textbf{OuroMamba-quantized W4A4 Zigma model exhibits the smallest FID increase}, maintaining high image generation quality. PTQ4VM, QMamba suffer significant degradation, under ultra low-precisions.

\begin{figure}[t]
  \centering \includegraphics[width = 0.80\columnwidth, keepaspectratio]{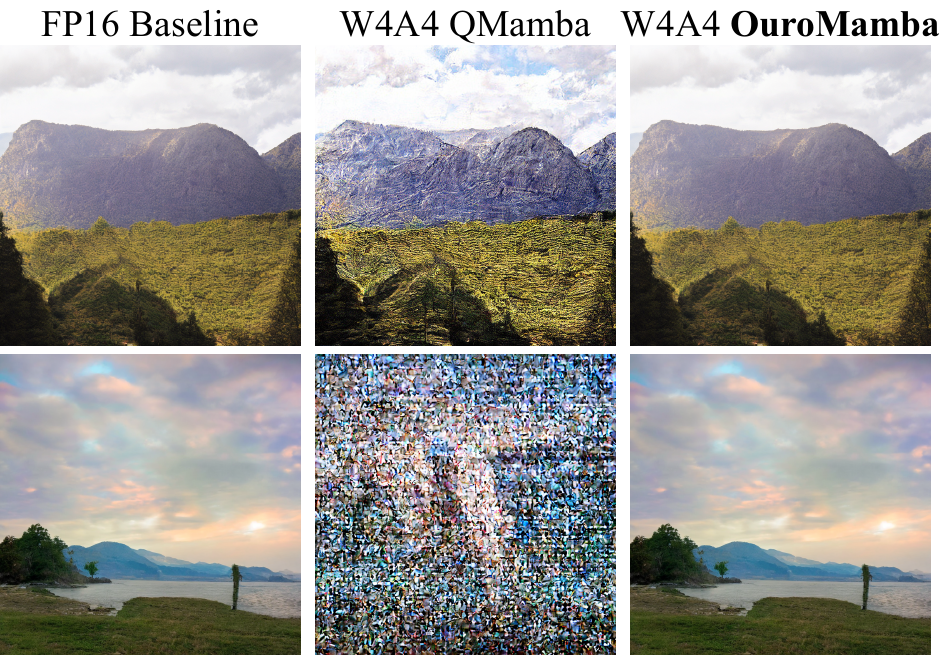}
  \vspace{-3mm}
  \caption{Visualization of landscape \cite{2021landscape} image generation by Zigma under different W4A4 quantization techniques.}
  \label{fig:zigma}
  \vspace{-5mm}
\end{figure}

\noindent
\textbf{Qualitative Analysis. }In \cref{fig:zigma}, we qualitatively compare the generated image quality of OuroMamba with the second best technique from \autoref{tab:diffusion}, QMamba, for the landscape image generation quality \cite{2021landscape}. As observed in \cref{fig:zigma}, \textbf{OuroMamba generates images indistinguishable from the FP16 baseline}, whereas QMamba produces significant noise artifacts and/or unreadable content. 

\begin{figure}[t]
  \centering \includegraphics[width = 0.75\columnwidth, keepaspectratio]{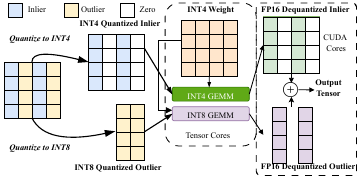}  
  \vspace{-2mm}
  \caption{Overview of W4A4 hybrid GEMM kernel.}
  \vspace{-3mm}
  \label{fig:gemm}
\end{figure}

\begin{figure}[t]
  \centering \includegraphics[width = \columnwidth, keepaspectratio]{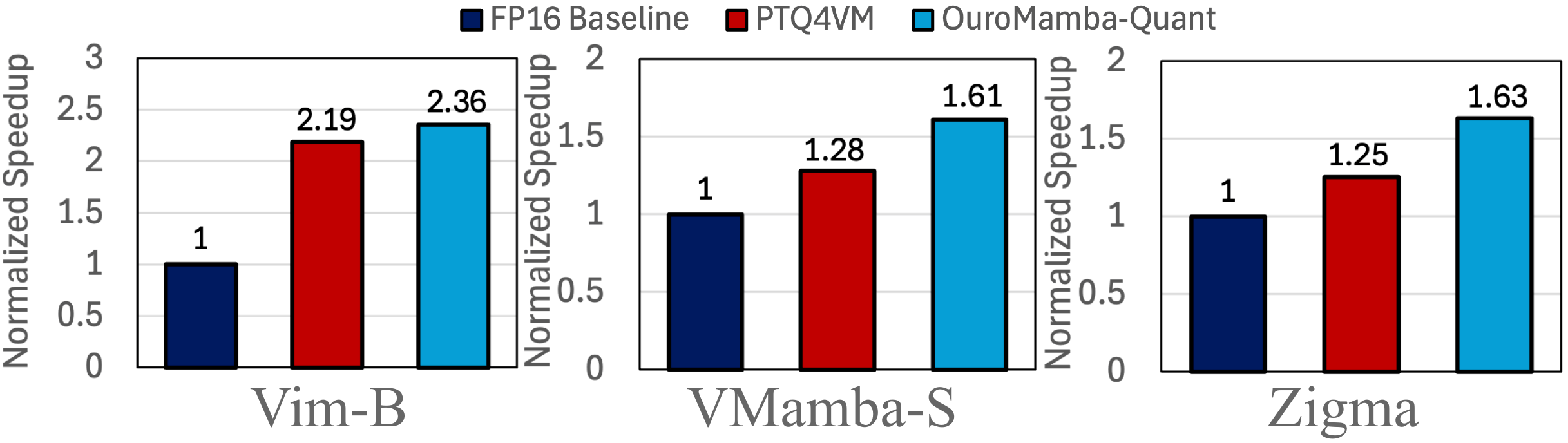}
  \vspace{-5mm}
  \caption{End-to-end latency speedup over FP16 baseline}
  \label{fig:speedup}
  \vspace{-6mm}
\end{figure}

\begin{figure*}[t]
  \centering \includegraphics[width = \linewidth, keepaspectratio]{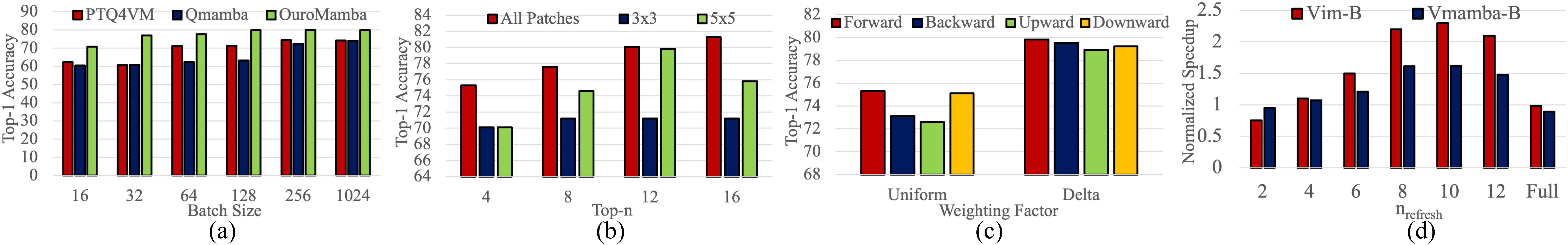}
  \vspace{-7mm}
  \caption{OuroMamba ablations for (a) Effect of batch size $\mathcal{B}$, (b) Effect of neighborhood size $\mathcal{N}$ and Top-$n$ positive patches, (c) Impact of weighting factor and scanning direction, (d) $n_{\texttt{refresh}}$ period.}
  \label{fig:ablation}
  \vspace{-6mm}
\end{figure*}

\subsection{GEMM Implementation and Evaluation}
\noindent
\textbf{Implementation. }Efficient inference of OuroMamba in VMMs requires addressing dynamic outlier extraction for real-time identification and mixed-precision GEMM operation for optimized computation. To this end, we implement the OuroMamba-\texttt{Quant} kernel using CUTLASS \cite{cutlass}. For efficient outlier channel extraction, activations are partitioned across channels and mapped to the thread blocks, where each block independently compares its assigned channels against the threshold $\theta$ to identify outliers and updates the $O_{\texttt{list}}$. To efficiently map the mixed-precision computation, we introduce a hybrid GEMM kernel. For a specific scenario of 4-bit weights with inlier and outlier activations of 4-bit and 8-bit, respectively, the hybrid GEMM executes one $4b\times 4b$ GEMM for inliers and one mixed-precision $4b\times 8b$ GEMM for outliers, as shown in \cref{fig:gemm}. We pack two consecutive 4-bit inlier activations into one byte, with outlier positions set to zero in the inlier buffer, and leverage the INT4 tensor cores for inlier GEMM. The 8-bit outlier channels are extracted and stored \textit{compactly} in a separate contiguous outlier buffer. The outliers execute GEMM on INT8 tensor cores performing computations only on the combined tensor of extracted outlier columns. Finally, the outputs from the inlier and outlier GEMMs are dequantized to FP16 and summed together, with all steps fused into a single pipeline for efficiency.

\noindent
\textbf{Evaluation. }As shown in \cref{fig:speedup}, OuroMamba-\texttt{Quant} achieves {speedup of} $\mathbf{2.36}\times$ \textbf{on Vim-B, } $\mathbf{1.61}\times$ \textbf{on VMamba-S} \textbf{and} $\mathbf{1.63}\times$ \textbf{on Zigma} over FP16 baselines with a $\mathcal{B}=128$. The speedup primarily stems from optimizations performed for efficient outlier extraction, mixed-precision GEMM through the INT4 and INT8 tensor cores. Unlike PTQ4VM, that uses a standalone kernel for dequantization, \textit{we fuse dequantization directly into the GEMM pipeline, eliminating associated overhead and improving latency}. Please refer supplementary for additional details.     

\begin{table}[t]
    \centering
    \label{tab:hyper_loss}
    \renewcommand*{\arraystretch}{1.0}
    \begin{minipage}[t]{0.45\columnwidth}
        \centering
        \caption{Hyperparameters used in evaluation.} 
        \vspace{-2mm}
        \label{tab:hyperparameters}
        \resizebox{\linewidth}{!}{%
        \begin{tabular}{lcc}
            \Xhline{2\arrayrulewidth}
            Parameter & Description & Value \\
            \Xhline{2\arrayrulewidth}
            $G$ & Gen. iterations & 1000 \\
            $\mathcal{N}$ & Neighborhood size & $5 \times 5$ \\
            $n$ & \# Positive patches &  12 \\
            $\mathcal{B}$ & Batch size & 128 \\
            $n_\texttt{ref.}$ & Refresh period & 10 \\
            $b_w$ & Weight precision & 4 \\
            $b_a^I$ & Inlier act. precision & 4,8 \\
            $b_a^O$ & Outlier act. precision & 8 \\
            \Xhline{2\arrayrulewidth}
        \end{tabular}
        \vspace{-2mm}
        }
    \end{minipage}\hfill 
    \begin{minipage}[t]{0.50\columnwidth}
        \centering
        \caption{Impact of different loss components for data generation.} 
        \vspace{-2mm}
        \label{tab:loss_func}
        \resizebox{\linewidth}{!}{%
        \begin{tabular}{ccccc}
            \Xhline{2\arrayrulewidth}
            $\mathcal{L}^{PSE}$ & $\mathcal{L}^{C}$ & $\mathcal{L}^{O}$ & W/A & Top-1 Acc. (\%)  \\
            \Xhline{2\arrayrulewidth}
             - & - & - & 32/32 & 81.60 \\ \cline{1-5}
             \cmark & \xmark & \xmark & 4/8 & 71.68 \\
             \xmark & \xmark & \cmark & 4/8 & 21.65 \\
             \cmark & \xmark & \cmark & 4/8 & 73.45 \\
             \xmark & \cmark & \xmark & 4/8 & 75.52 \\ 
             \xmark & \cmark & \cmark & 4/8 & \textbf{79.81} \\
             \cmark & \cmark & \cmark & 4/8 & 74.32 \\
            \Xhline{2\arrayrulewidth}
        \end{tabular}
        \vspace{-2mm}
        }
    \end{minipage}
    \vspace{-5mm}
\end{table}

\subsection{Discussions and Ablations}
\noindent
\textbf{Effect of Batch Size $\mathcal{B}$. }We present the accuracy comparison across batch sizes from 16 to 1024 in \cref{fig:ablation}(a) for W4A8 Vim-S on image classification. OuroMamba exhibits minimal accuracy gains beyond 128, validating this as the optimal batch size. In contrast, VMM PTQ techniques PTQ4VM and QMamba rely on significantly larger batch sizes of 256 and 1024 real-data samples, respectively. Notably, even at batch sizes below 128, OuroMamba consistently outperforms PTQ4VM and QMamba, highlighting the effectiveness of its synthetic data generation which is tailored for quantization of the respective model.   

\noindent
\textbf{Effect of Neighborhood Size $\mathcal{N}$ and Top-$n$ Patches. } \cref{fig:ablation}(b) demonstrates the effect of different neighborhood sizes $\mathcal{N}$ and top-$n$ positive patch selection on the quality of synthetic data generated by {OuroMamba}-\texttt{Gen}. To assess this, we measure the {top-1 accuracy of W4A8 Vim-S} using the generated samples are as for calibration. Notably, a $5\times5$ neighborhood size with the top-12 positive patches yields the highest accuracy. This choice of $\mathcal{N}$ effectively captures complex spatial dependencies while remaining computationally efficient compared to using all patches in $\alpha_p$.

\noindent
\textbf{Impact of Scanning Direction. }We generate synthetic data by applying it independently to each scanning direction of VMamba-B. Using the same evaluation setup, we measure the {top-1 accuracy for W4A4 quantization}. As shown in Fig. \cref{fig:ablation}(c), under delta weighting, we observe that regardless of the scanning direction, the resulting model achieves similar accuracy. 
 
\noindent
\textbf{Impact of Weighting Factor.} \cref{fig:ablation}(c) shows the effect of uniform weighting compared to the $\Delta(t)$ weighting on the synthetic data generation by {OuroMamba}-\texttt{Gen}. For W4A4 VMamba-B, following the same evaluation setup as above, we see independent of different scanning directions {delta weighting consistently yields accuracy improvements}.

\noindent
\textbf{Choice of Objective Function. }The study in \autoref{tab:loss_func} evaluates the impact of different loss function components in $\mathcal{L}^{gen}$ on synthetic data generation effectiveness and their influence on the {top-1 accuracy of W4A8 quantization} for Vim-S. We empirically validate our choice of \textit{patch-level contrastive learning} ($\mathcal{L}^{C}$) over the patch-similarity metric ($\mathcal{L}^{PSE}$) used in prior ViT DFQ techniques \cite{li2022patch, li2023psaq}, demonstrating its superior effectiveness in generating high-quality calibration data. As shown in \autoref{tab:loss_func}, a linear combination of $\mathcal{L}^{C}$ and $\mathcal{L}^{O}$ yields the highest accuracy, while using only $\mathcal{L}^{O}$ results in the lowest, highlighting its limited role in capturing semantic relationships for synthetic data generation. Although $\mathcal{L}^{PSE}$ combined with $\mathcal{L}^{O}$ achieves acceptable accuracy, it still underperforms compared to $\mathcal{L}^{gen}$.

\noindent
\textbf{Effect of $n_{\texttt{refresh}}$. }In \cref{fig:ablation}(d), we present the normalized speedup for W4A4 Vim-B and VMamba-B across different values of the periodic refresh parameter $n_{\texttt{refresh}}$. We observe that $n_{\texttt{refresh}} = 10$ time-steps yields the highest speedup. Reducing $n_{\texttt{refresh}}$ leads to a sharp decline due to frequent reinitialization of $O_{\texttt{list}}$ and repeated outlier extraction over the entire tensor. Conversely, increasing $n_{\texttt{refresh}}$ beyond 10 results in excessive outlier accumulation, increasing computational complexity and reducing speedup. Notably, when $n_{\texttt{refresh}} = {Full}$ (i.e., no refresh), {OuroMamba-\texttt{Quant} becomes slower than the FP16 baseline} due to the overhead of processing accumulated outliers.

\noindent
\textbf{Limitation Discussion. }The dynamic outlier selection assumes that inlier distributions remain stable relative to their values determined during calibration. In our experiments across all models, we observed no significant variations in the dynamic range of inliers. However, if future model architectures exhibit substantial fluctuations in inlier values, additional investigation will be necessary.

\section{Conclusions}
\label{sec:conc}
This paper presents OuroMamba, a two-stage DFQ framework that addresses key challenges in enabling DFQ for VMMs. OuroMamba-\texttt{Gen} enhances implicit attention through patched neighborhood interactions and applies patch-level contrastive learning to generate semantically meaningful synthetic data. OuroMamba-\texttt{Quant} employs mixed-precision quantization with dynamic outlier detection, minimizing quantization error. Extensive evaluations demonstrate the SoTA quantization performance and speedup of our custom kernel across diverse vision tasks.

\section*{Acknowledgments}
This work was supported in part by CoCoSys, one of the seven centers in JUMP 2.0, a Semiconductor Research Corporation (SRC) program sponsored by DARPA.


{
    \small
    \bibliographystyle{ieeenat_fullname}
    \bibliography{refs}
}


\end{document}